\documentclass[conference]{IEEEtran}
\IEEEoverridecommandlockouts
\usepackage{cite}
\usepackage{algorithm,amsmath,amssymb,amsfonts}
\usepackage{algpseudocode}
\usepackage{graphicx}
\usepackage{textcomp}
\usepackage{xcolor}
\def\BibTeX{{\rm B\kern-.05em{\sc i\kern-.025em b}\kern-.08em
    T\kern-.1667em\lower.7ex\hbox{E}\kern-.125emX}}
\begin{document}

\title{\huge Deep Reinforcement Learning  for Time Scheduling in RF-Powered Backscatter Cognitive Radio Networks\\
}
\author{\IEEEauthorblockN{Tran The Anh$^1$, Nguyen Cong Luong$^1$,  Dusit Niyato$^1$, Ying-Chang Liang$^2$, and Dong In Kim$^3$}
\IEEEauthorblockA{\textit{}
$^1$School of Computer Science and Engineering, Nanyang Technological University, Singapore}
$^2$Center for Intelligent Networking and Communications, University of Electronic Science and Technology of China, China\\
$^3$School of Information and Communication Engineering, Sungkyunkwan University, Korea
}

\maketitle

\begin{abstract}
In an RF-powered backscatter cognitive radio network, multiple secondary users communicate with a secondary gateway by backscattering or harvesting energy and actively transmitting their data depending on the primary channel state. To coordinate the transmission of multiple secondary transmitters, the secondary gateway needs to schedule the backscattering time, energy harvesting time, and transmission time among them. However, under the dynamics of the primary channel and the uncertainty of the energy state of the secondary transmitters, it is challenging for the gateway to find a time scheduling mechanism which maximizes the total throughput. In this paper, we propose to use the deep reinforcement learning algorithm to derive an optimal time scheduling policy for the gateway. Specifically, to deal with the problem with large state and action spaces, we adopt a Double Deep-Q Network (DDQN) that enables the gateway to learn the optimal policy. The simulation results clearly show that the proposed deep reinforcement learning algorithm outperforms non-learning schemes in terms of network throughput. 
\end{abstract}


\section{Introduction}
Radio Frequency (RF)-powered cognitive radio networks are considered to be a promising solution which improves radio spectrum utilization and efficiency as well as addresses the energy constraint issue for low-power secondary systems, e.g., the IoT system~\cite{huynh2018}, \cite{li2018}, \cite{kang2018}. However, in the RF-powered cognitive radio networks, RF-powered secondary transmitters typically require a long time period to harvest sufficient energy for their active transmissions. This may significantly deteriorate the network performance. Thus, RF-powered cognitive radio networks with ambient backscatter\cite{liu2013} have been recently proposed. In the RF-powered backscatter cognitive radio network, a primary transmitter, e.g., a base station, transmits RF signals on a licensed channel. When the channel is busy, the secondary transmitters either transmit their data to a secondary gateway by using backscatter communications or harvest energy from the RF signals through RF energy harvesting techniques. When the channel is idle, the secondary transmitters use the harvested energy to transmit their data to the gateway. As such, the RF-powered backscatter cognitive radio network enables secondary systems to simultaneously optimize the spectrum usage and energy harvesting
to maximize their performance. However, one major problem in the RF-powered backscatter cognitive radio network is how the secondary gateway\footnote{We use ``secondary gateway'' and ``gateway'' interchangeably in the paper.} schedules the backscattering time, energy harvesting time, and transmission time among multiple secondary transmitters so as to maximize the network throughput. 

To address the problem, optimization methods and game theory can be used. The authors in \cite{wang2018} optimized the time scheduling for the gateway in the RF-powered backscatter cognitive radio network through using the Stackelberg game. In the game, the gateway is the leader, and the secondary transmitters are the followers. The gateway first determines spectrum sensing time and an interference price to maximize its revenue. Based on the time and price, each secondary transmitter determines the energy harvesting time, backscattering time, and transmission time so as to maximize its throughput. However, the proposed game requires complete and perfect sensing probability information, and thus the game cannot deal with the dynamics of the network environment.



To optimize the performance of RF powered backscatter cognitive radio in the dynamic environment and with large state and action space, deep reinforcement learning (DRL) technique~\cite{mnih2015} can be adopted. In principle, the DRL implements a Deep Q-Network (DQN), i.e., the combination of a deep neural network and the Q-learning, to derive an approximate value of Q-values of actions, i.e., decisions. Compared with the conventional reinforcement learning, the DRL can improve significantly the learning performance and the learning speed. Therefore, in this paper, we propose to use the DRL for the time scheduling in the RF-powered backscatter cognitive radio network. In particular, we first formulate a stochastic optimization problem that maximizes the total throughput for the network. The DRL algorithm is adopted to achieve the optimal time scheduling policy for the secondary transmitters. To overcome the instability of the learning and to reduce the overestimation of action values, the Double DQN (DDQN) is used to implement the DRL algorithm. Simulation results show that the proposed DRL algorithm always achieves the better performance compared with non-learning algorithms. To the best of our knowledge, this is the first paper that investigates an application of DRL in the RF-powered backscatter cognitive radio network. 

The rest of this paper is organized as follows. Section II reviews
related work. Section III describes the system model and
problem formulation. Section IV presents the DRL algorithm for the time scheduling in the RF-powered backscatter cognitive radio network. Section V shows the performance evaluation
results. Section VI summarizes the paper.

\section{Related Work}

Backscatter communications systems can be optimized to achieve optimal throughput. In~\cite{hoang2017}, the authors considered the data scheduling and admission control problem of a backscatter sensor network. The authors formulated the problem as a Markov decision process, and learning algorithm was applied to obtain the optimal policy that minimizes the weighted sum of delay of different types of data. In~\cite{lyu2018a}, the authors formulated an optimization problem for ambient backscatter communications networks. The problem aims to derive an optimal control policy for sleep and active mode switching and reflection coefficient used in the active mode. However, only single transmitter was considered. In~\cite{lyu2018b}, the authors extended the study in~\cite{lyu2018a} to a cognitive radio network. Specifically, the hybrid HTT and backscatter communications are adopted and integrated for a secondary user. The authors proposed an optimal time allocation scheme which is based on the convex optimization problem. While multiple secondary users were considered, the secondary user does not employ energy storage. The authors in~\cite{yang2018} analyzed the backscatter wireless powered communication system with multiantenna by using the stochastic geometry approach. The energy and information outage probabilities in the energy harvesting and backscatter were derived. Then, the authors introduced an optimization problem for time allocation to maximize overall network throughput. The authors in~\cite{kwan2018} introduced a two-way communication protocol for backscatter communications. The protocol combines time-switching and power splitting receiver structures with backscatter communication. The optimization problem was formulated and solved to maximize sum throughput of multiple nodes. Unlike the above works that consider time allocation, the authors in~\cite{gong2018} proposed a channel-aware rate adaptation protocol for backscatter networks. The protocol first probes the channel and adjusts the transmission rate of the backscatter transmitter. The objective is to minimize the number of channels to be used for successfully communicating with all backscatter nodes.

Although a few works in the literature studied the performance optimization of backscatter-based communications networks, almost all of them assume that information about the network is always available which may not be realistic under random and unpredictable wireless environments. Therefore, this paper considers a scenario in which the network does not have complete information. The network needs to learn to assign backscattering time, energy harvesting time, and transmission time to the secondary transmitters to maximize the total network throughput.

\section{System Model}
\begin{figure}[h]
 \centering
\includegraphics[width=7.7cm, height = 5.6cm]{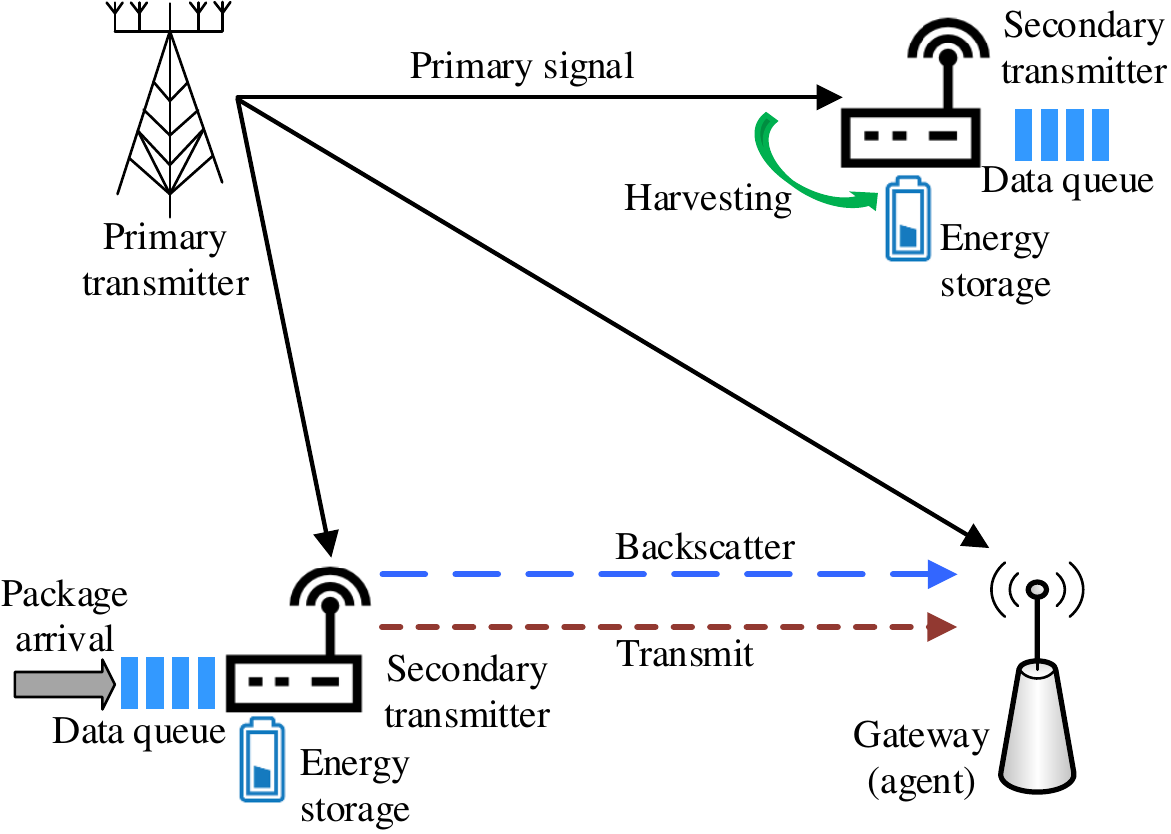}
 \caption{Ambient backscatter model with multiple secondary transmitters.}
  \label{backscatter_model}
\end{figure}

\begin{figure}[h]
 \centering
\includegraphics[width=\linewidth]{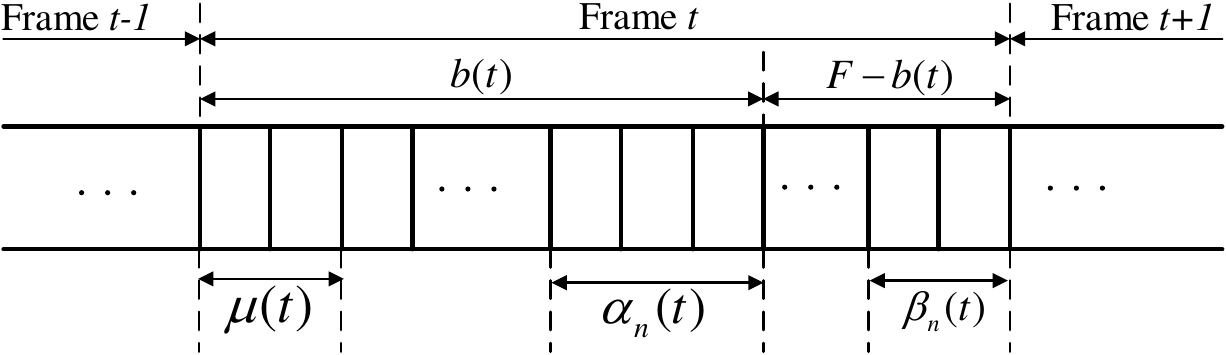}
 \caption{Structure of frame $t$ and time scheduling.}
  \label{Timeframe}
\end{figure}

We consider the RF-powered backscatter cognitive radio network. The network consists of a primary transmitter and $N$ secondary transmitters (Fig.~\ref{backscatter_model}). The primary transmitter transmits RF signal on a licensed channel. The transmission is organized into frames, and each frame is composed of $F$ time slots. In Frame $t$ (see Fig.~\ref{Timeframe}), the transmission duration of the primary transmitter or the busy channel period, i.e., the number of time slots the channel is busy, is denoted by $b(t)$ which is random. The secondary transmitters transmit data to the gateway. The gateway also controls the transmission scheduling of the secondary transmitters. In particular, during the busy channel period, the time slots can be assigned for energy harvesting by all the secondary transmitters. The number of time slots for energy harvesting is denoted by $\mu(t)$. Let $e^{\mathrm{h}}_n$ denote the number of energy units that secondary transmitter $n$ harvests in one busy time slot. The harvested energy is stored in energy storage, e.g., a super-capacitor, of the secondary transmitter, the maximum capacity of which is denoted by $C_n$. A secondary transmitter has a data queue which stores an incoming packet, e.g., from its sensor device. The maximum capacity of the data queue is denoted by $Q_n$, and the probability that a packet arrives at the queue in a time slot is denoted by $\lambda_n$. Then, the rest of the busy time slots, i.e., $b(t)-\mu(t)$, will be allocated for the secondary transmitters to transmit their data by using backscatter communications, i.e., the {\em backscatter mode}. The number of time slots for backscatter transmission by secondary transmitter $n$ is denoted by $\alpha_n(t)$, and the number of packets transmitted in each time slot is $d^{\mathrm{b}}_n$. Then, during the idle channel period which has $F-b(t)$ time slots, the secondary transmitters can transfer data to the gateway by using active-RF transmission, i.e., the {\em active mode}. The number of time slots for secondary transmitter $n$ to transmit in the active mode is denoted by $\beta_n(t)$. Each time slot can be used to transmit $d^{\mathrm{a}}_n$ packets from the data queue, and the secondary transmitter consumes $e^{\mathrm{a}}_n$ units of energy from the storage. The data transmission in the backscatter mode and in the active mode is successful with the probabilities $S^{\mathrm{b}}_n$ and $S^{\mathrm{a}}_n$, respectively. Hence, the total throughput of the RF-powered backscatter cognitive radio network is the sum of the number of packets successfully transmitted by all secondary transmitters.

\section{Problem Formulation}

To optimize the total throughput, we formulate a stochastic optimization problem for the RF-powered backscatter cognitive radio network. The problem is defined by a tuple $<{\mathcal{S}}, {\mathcal{A}}, {\mathcal{P}}, {\mathcal{R}}>$.
\begin{itemize}
	\item ${\mathcal{S}}$ is the state space of the network.
	\item ${\mathcal{A}}$ is the action space.
	\item ${\mathcal{P}}$ is the state transition probability function, where $P_{s,s'}(a)$ is the probability that the current state $s \in {\mathcal{S}}$ transits to the next state $s' \in {\mathcal{S}}$ when action $a \in {\mathcal{A}}$ is taken.
	\item ${\mathcal{R}}$ is the reward function of the network.
\end{itemize} 

The state space of secondary transmitter $n$ is denoted by 
\begin{equation}
	{\mathcal{S}}_n	=	\Big\{ (q_n, c_n) ; q_n \in \{0,1,\ldots,Q_n\}, c_n \in \{0,1,\ldots,C_n\} \Big\},
\end{equation}
where $q_n$ represents the queue state, i.e., the number of packets in the data queue, and $c_n$ represents the energy state, i.e., the number of energy units in the energy storage. Let the channel state, i.e., the number of busy time slots, be denoted by ${\mathcal{S}}^{\mathrm{c}} = \{ (b); b \in \{0,1,\ldots,F \} \}$. Then, the state space of the network is defined by
\begin{equation}
	{\mathcal{S}}	=	{\mathcal{S}}^{\mathrm{c}} \times \prod_{n = 1}^N {\mathcal{S}}_n	,
\end{equation}
where $\times$ and $\prod$ represent the Cartesian product.

The action space of the network is defined as follows:
\begin{eqnarray}
	{\mathcal{A}}	& = & 	\Bigg\{ (\mu, \alpha_1,\ldots,\alpha_N, \beta_1,\ldots,\beta_N ); \nonumber	\\
					& & \mu + \sum_{n=1}^N  \alpha_n \leq b, \mu + \sum_{n=1}^N ( \alpha_n + \beta_n ) \leq F \Bigg\},
\label{eq:actionspace}
\end{eqnarray}
where again $\mu$ is the number of busy time slots that are used for energy harvesting by the secondary transmitters, $\alpha_n$ is the number of busy time slots that secondary transmitter $n$ transmits data in the backscatter mode, and $\beta_n$ is the number of idle time slots that secondary transmitters $n$ transmits data in the active mode. The constraint $\mu + \sum_{n=1}^N  \alpha_n \leq b$ ensures that the number of time slots for energy harvesting and all backscatter transmissions do not exceed the number of busy time slots. Likewise, the constraint $\mu + \sum_{n=1}^N ( \alpha_n + \beta_n ) \leq F$ ensures that the number of time slots for energy harvesting, all transmissions in the backscatter and active modes do not exceed the total number of time slots in a frame.

Now, we consider the state transition of the network. In the busy channel period, the number of time slots assigned to secondary transmitter $n$ for harvesting energy is $b(t) - \alpha_n$. Thus, after the busy channel period, the number of energy units in the storage of the secondary transmitter changes from $c_n$ to $c_n^{(1)}$ as follows:
\begin{equation}
	c_n^{(1)}=\min \big(c_n + (b(t) - \alpha_n)e^{\mathrm{h}}_n,C_n \big).
\end{equation}

Also, the number of packets in the data queue of secondary transmitter $n$ changes from $q_n$ to $q_n^{(1)}$ as follows:
\begin{equation}
	q_n^{(1)}	=	\max \big(0, q_n - \alpha_n d^{\mathrm{b}}_n \big).
\end{equation}

In the idle channel period, secondary transmitter $n$ requires $q_n^{(1)}/d^{\mathrm{a}}_n$ time slots to transmit $q_n^{(1)}$ packets. However, the secondary transmitter is only assigned with $\beta_n$ time slots for the data transmission. Thus, it actually transmits its packets in $\min (\beta_n, q_n^{(1)}/d^{\mathrm{a}}_n)$ time slots. 

After the idle channel period, the energy state of secondary transmitter $n$ changes from $c_n^{(1)}$ to  $c'_n$ as follows:
\begin{equation}
	c'_n	=	\max \big(0, c_n^{(1)} - \min (\beta_n, q_n^{(1)}/d^{\mathrm{a}}_n)e^{\mathrm{a}}_n \big).
\end{equation}

Also, the number of packets in the data queue of secondary transmitter $n$ changes from $q_n^{(1)}$ to $q_n^{(2)}$ as follows:
\begin{equation}
	q_n^{(2)}	=	\max \big(0, q_n^{(1)} - \min (\beta_n, c_n^{(1)}/e^{\mathrm{a}}_n)d^{\mathrm{a}}_n \big).
\end{equation}

Note that new packets can arrive at each time slot with a probability of $\lambda_n$. We assume that the new packets are only added to the data queue when the time frame finishes. Thus, at the end of the time frame, the number of packets in the data queue of secondary transmitter $n$ changes from $q_n^{(2)}$ to $q'_n$ as follows:
\begin{equation}
	q'_n	=	q_n^{(2)} + p_n,
\end{equation}
where $p_n$ is the number of packets arriving in the secondary transmitter during the time frame. $p_n$ typically follows binomial distribution $B(F,\lambda _n)$~\cite{bliss1953}. Then, the probability of $m$ packets arriving in the secondary transmitter during $F$ time slots is 
\begin{equation}
Pr(p_n = m )=\binom{F}{m} \lambda_n^m(1-\lambda_n)^{F-m}.
\end{equation}

The reward of the network is defined as a function of state $s \in {\mathcal{S}}$ and action $a \in {\mathcal{A}}$ as follows:
\begin{equation}
	\mathcal{R}(s,a)=	\sum_{n=1}^N S^{\mathrm{b}}_n (q_n^{(1)} - q_n) + \sum_{n=1}^N S^{\mathrm{a}}_n (q_n^{(2)} - q_n^{(1)}).
	\label{eq:reward}
\end{equation}

The first and the second terms of the reward expression in (\ref{eq:reward}) are for the total numbers of packets transmitted in the backscatter and active modes, respectively. 

To obtain the mapping from a network state $s \in {\mathcal{S}}$ to an action $a \in {\mathcal{A}}$ such that the accumulated reward is maximized, conventional algorithm can be applied. The goal of the algorithm is to obtain the optimal policy defined as $\pi^* :  {\mathcal{S}} \rightarrow  {\mathcal{A}}$. In the algorithm, the optimal policy to maximize value-state function is defined as follows:
\begin{equation}
	V(s)  = 	\mathbb{E} \left[	\sum_{t=0}^{T-1}	\gamma^t	{\mathcal{R}}(s(t), a(t) )	\right]	,
\end{equation}
where $T$ is the length of the time horizon, $\gamma$ is the discount factor for $0\leq \gamma < 1$, and $\mathbb{E}[\cdot]$ is the expectation. Here, we define $a = \pi(s)$ which is the action taken at state $s$ given the policy $\pi$. With the Markov property, the value function can be expressed as follows:
\begin{eqnarray}
	V(s)	& = &  	\sum_{s' \in {\mathcal{S}} } P_{\pi(s) } (s, s') \left( {\mathcal{R}}(s,a) + \gamma V(s') \right)	,	\\
	\pi(s)	& = & 	\max_{ a \in {\mathcal{A}} } \left(	\sum_{s' \in {\mathcal{S}} } P_{\pi(s) } (s, s') \Big( {\mathcal{R}}(s,a) + \gamma V(s') \Big)	\right)	.
\end{eqnarray}
With the Q-learning algorithm, Q-value is defined, and its optimum can be obtained from the Bellman's equation, which is given as
\begin{equation}
	Q(s,a)	=  	\sum_{s' \in {\mathcal{S}} } P_{\pi(s) } (s, s') \left( {\mathcal{R}}(s,a) + \gamma V(s') \right)	.
\end{equation}
The Q-value is updated as follows:
\begin{align}
\label{Q_value_update}
	Q^{\mathrm{new}}(s,a)	= &  	(1-l) Q(s,a) 	\\
				& + l \left( r(s,a) + \gamma \max_{a' \in {\mathcal{A}} }  Q( s', a') \right).\notag
\end{align}
where $l$ is the learning rate, and $r(s,a)$ is the reward received.

However, the standard algorithms and Q-learning to solve the stochastic optimization problem all suffer from large state and action space of the networks. Thus, we resort to the deep reinforcement learning algorithm.

\section{Deep Reinforcement Learning Algorithm}
By using (\ref{Q_value_update}) to update Q-values in a look-up table, the Q-learning algorithm can efficiently solve the optimization problem if the state and action spaces are small. In particular for our problem, the gateway needs to observe states of all $N$ secondary transmitters and choose actions from their action spaces. As $N$ grows, the state and action spaces can become intractably large, and several Q-values in the table may not be updated. To solve the issue, we propose to use a DRL algorithm.

Similar to the Q-learning algorithm, the DRL allows the gateway to map its state to an optimal action. However, instead of using the look-up table, the DRL uses a Deep Q-Network (DQN), i.e.,  a multi-layer neural network with weights $\boldsymbol{\theta}$, to derive an approximate value of $Q^*(s,a)$. The input of the DQN is one of the states of the gateway, and the output includes Q-values $Q(s,a;\boldsymbol{\theta})$ of all its possible actions. To achieve the approximate value $Q^*(s,a)$, the DQN needs to be trained by using transaction $< s,a,r,s'>$, i.e., experience, in which action $a$ is selected through using the $\epsilon$-greedy policy. Training the DQN is to update its weights $\boldsymbol{\theta}$ to minimize a loss function defined as:
\begin{equation}
L= \mathbb{E}\left[ (y-Q(s,a;\boldsymbol{\theta}))^2\right], 
\label{DQN_loss}
\end{equation}
where $y$ is the target value. $y$ is given by
\begin{equation}
y= r+ \gamma \max_{a' \in {\mathcal{A}} }  Q( s', a';\boldsymbol{\theta^-}),
\label{DQN_y_value}
\end{equation}
where $\boldsymbol{\theta^-}$ are the old weights, i.e., the weights from the last iteration, of the DQN. 

Note that the $\max$ operator in (\ref{DQN_y_value}) uses the same Q-values both to select and to evaluate an action of the gateway. This means that the same Q-values are being used to decide which action is the best, i.e., the highest expected reward, and they are also being used to estimate the action value. Thus, the Q-value of the action may be over-optimistically estimated which reduces the network performance. 

To prevent the overoptimism problem, the action selection should be decoupled from the action evaluation~\cite{van2016}. Therefore, we use the Double DQN (DDQN). The DDQN includes two neural networks, i.e., an online network with weights $\boldsymbol{\theta}$ and a target network with weights $\boldsymbol{\theta^-}$. The target network is the same as the online network that its weights $\boldsymbol{\theta^-}$ are reset to $\boldsymbol{\theta}$ of the online network every $L^-$ iterations. At other iterations, the weights of the target network keep unchanged while those of the online network are updated at each iteration. 

In principle, the online network is trained by updating its weights $\boldsymbol{\theta}$ to minimize the loss function as shown in~(\ref{DQN_loss}). However, $y$ is replaced by $y^{DDQN}$ defined as
\begin{equation}
y^{DDQN}=r + \gamma Q\Big{(} s', \arg\max_{a' \in \mathcal{A}} Q_i(s',a';\boldsymbol{\theta});\boldsymbol{\theta}^{-}\Big{)}.
\label{DQN_y_value_DDQN}
\end{equation}

As shown in (\ref{DQN_y_value_DDQN}), the action selection is based on the current weights $\boldsymbol{\theta}$, i.e., the weights of the online network. The weights $\boldsymbol{\theta^{-}}$ of the target network are used to fairly evaluate the value of the action.

The DDQN algorithm for the gateway to find its optimal policy is shown in Algorithm~\ref{DDQN_Algorithm}. Accordingly, both the online and target networks use the next state $s'$ to compute the optimal value $Q(s',a';\boldsymbol{\theta})$. Given the discount factor $\gamma$ and the current reward $r$, the target value $y^{DDQN}$ is obtained from (\ref{DQN_y_value_DDQN}). Then, the loss function is calculated as defined in~(\ref{DQN_loss}) in which $y$ is replaced by $y^{DDQN}$. The value of the loss function is back propagated to the online network to update its weights $\boldsymbol{\theta}$. Note that to address the instability of the learning of the algorithm, we adopt an experience replay memory $\mathcal{D}$ along with the DDQN. As such, instead of using the most recent transition, a random mini-batch of transactions is taken from the replay memory to train the Q-network. 
\begin{algorithm}
\small
\caption{DDQN algorithm for time scheduling of the gateway.}\label{DDQN_Algorithm}

\hspace*{\algorithmicindent} \textbf{Input:} Action space $\mathcal{A}$; mini-batch size $L_b$; target network replacement frequency $L^{-}$.\\

\hspace*{\algorithmicindent} \textbf{Output:} Optimal policy $\pi^*$.

\begin{algorithmic}[1]

\State \textbf{Initialize:} Replay memory $\mathcal{D}$; online network weights $\boldsymbol{\theta}$; target network weights $\boldsymbol{\theta^-} = \boldsymbol{\theta}$; online action-value function $Q(s,a; \boldsymbol{\theta})$; target action-value function $Q({s}', {a}'; \boldsymbol{\theta^-})$; $k=i=0$.

\Repeat \text{ for each episode} $i$:

\State Initialize network state $s$ after receiving state massages from the primary transmitter and $N$ secondary transmitters.

\Repeat \text{ for each iteration $k$ in episode} $i$:

\State Choose action $a$ according to $\epsilon-greedy$ policy from $Q(s,a; \boldsymbol{\theta})$.

\State Broadcast time scheduling massages defined by $a$ to $N$ secondary transmitters.
\State Receive an immediate reward $r_k$.
\State Receive state massages from primary transmitter and $N$ secondary transmitters and update next network state $s'$.
\State Store tuple $(s, a, r_k, {s}')$ in $\mathcal{D}$.

\State Sample a mini-batch of $L_b$ tuples $(s, a, r_t, {s}')$ from $\mathcal{D}$.

\State Define $a^{\text{max}}=\arg \max_{a' \in \mathcal{A}} Q({s}', {a}'; \boldsymbol{\theta})$.

\State Determine

\begin{equation}
y_{t}^{DDQN}=\begin{cases}
r_t,\text{ if episode $i$ terminates at iteration } t + 1,\\
r_t+ \gamma Q\left ({s}', a^{\text{max}};\boldsymbol{\theta^-} \right), \text{ otherwise}.\notag\\
\end{cases}
\end{equation}


\State Update $\boldsymbol{\theta}$ by performing a gradient descent step on $(y_{t}^{DDQN}-Q(s,a;\boldsymbol{\theta}))^{2}$.

\State Reset $\boldsymbol{\theta^-} = \boldsymbol{\theta}$ every $L^{-}$ steps.

\State Set $s\leftarrow {s}'$.

\State Set $k=k+1$.

\Until {$k$ is greater than the maximum number of steps in episode $i$.}

\State Set $i=i+1$.

\Until {$i$ is greater than the desired number of episodes.}

\end{algorithmic}
\end{algorithm}


\section{Performance Evaluation}
In this section, we present experimental results to evaluate the performance of the proposed DRL algorithm. For comparison, we introduce the HTT~\cite{lyu2018b}, the backscatter communication~\cite{liu2013}, and a random policy as baseline schemes. In particular for the random policy, the gateway assigns time slots to the secondary transmitters for the energy harvesting, data backscatter, and data transmission, by choosing randomly a tuple $(\mu, \alpha_1,\ldots,\alpha_N, \beta_1,\ldots,\beta_N )$ in action space $\mathcal{A}$. Note that we do not introduce the reinforcement learning algorithm~\cite{huynh2018} since it cannot be run in our computation environment as the problem is too complex. The simulation parameters for the RF-powered backscatter cognitive radio network are shown in Table~\ref{table:parameters_CRN}, and those for the DRL algorithm are listed in Table~\ref{table:parameters_system}. The DRL algorithm is implemented by using the TensorFlow deep learning library. The Adam optimizer is used that allows to adjust the learning rate during the training phase. The $\epsilon$-greedy policy with $\epsilon=0.9$ is applied in the DRL algorithm to balance the exploration and exploitation. This means that a random action is selected with a probability of $\epsilon =0.9$, and the best action, i.e., the action that maximizes the Q-value, is selected with a probability of $\epsilon =0.1$. To move from a more explorative policy to a more exploitative one, the value of $\epsilon$ is linearly reduced from $0.9$ to $0$ during the training phase. 
\begin{table}[!h]
\caption{Backscatter cognitive radio network parameters}
\label{table:parameters_CRN}
\centering
\begin{tabular}{lc}
\hline\hline
{\em Parameters} 			& {\em Value} \\ [0.5ex]
\hline
Number of secondary transmitters               ($N$)    & 3             \\ 
Number of time slots in a time frame ($F$) & 10 \\ 
Number of idle time slots in a time frame ($b(t)$)& {[}1;9{]}    \\ 
Data queue size ($Q_n$)     & 10   \\ 
Energy storage capacity		           ($C_n$)    & 10   \\ 
Packet arrival probability     ($\lambda_n $)  & {[}0.1;0.9{]}        \\
$d^\mathrm{b}_n$  & 1 \\
$d^\mathrm{a}_n$  & 2 \\
$e^\mathrm{h}_n$  & 1 \\
$e^\mathrm{a}_n$ &  1 \\
\hline
\end{tabular}
\label{table:parameters}
\end{table}

\begin{table}[!h]
\caption{System model parameters}
\label{table:parameters_system}
\centering
\begin{tabular}{ll}
\hline\hline
{\em Parameters} 			& {\em Value} \\ [0.5ex]
\hline
Number of hidden layers             & 3                   \\
Fully connected neuron network size & 32x32x32            \\
Activation                          & ReLU                \\
Optimizer 			& Adam                \\
Learning rate             & 0.001               \\ 
Discount rate ($\gamma$)            & 0.9                 \\ 
$\epsilon$-greedy                   & 0.9 $\rightarrow$ 0 \\ 
Mini-batch size ($L_b$)                    & 32                  \\
Replay memory size		& 50000               \\ 
Number of iterations per episode         & 200                 \\
Number of training iterations              & 1000000             \\
Number of iterations for updating target network ($L^{-}$)& 10000               \\
\hline
\end{tabular}
\label{table:parameters}
\end{table}

\begin{figure}[!h]
 \centering
\includegraphics[width=6.7cm, height = 5.3cm]{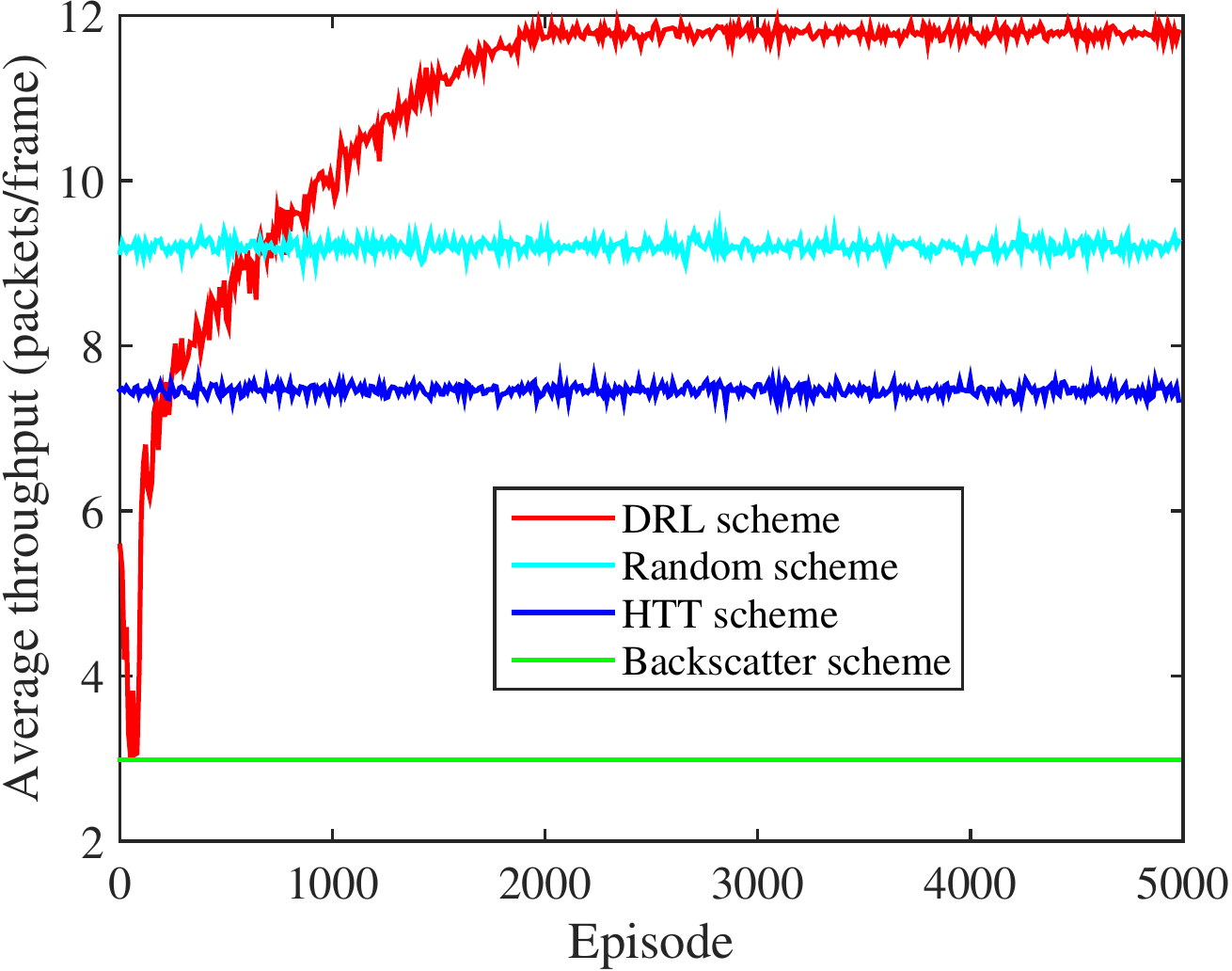}
 \caption{Average throughput comparison between the proposed DRL scheme and the baseline schemes.}
 \label{throughput_comparison}
\end{figure}

\begin{figure}[!h]
 \centering
 \includegraphics[width=6.7cm, height = 5.3cm]{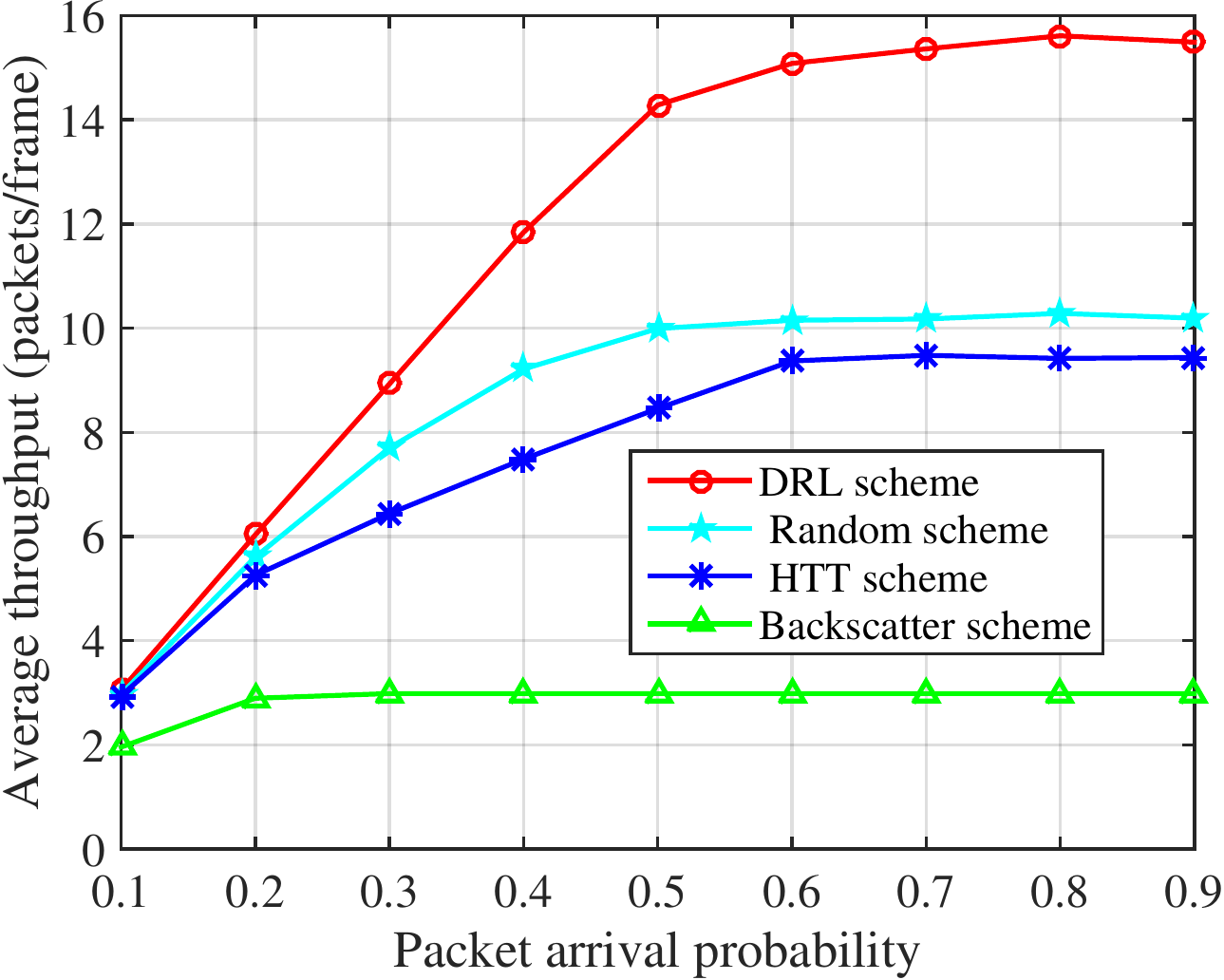}
 \caption{Average throughput versus packet arrival probability.}
  \label{data_rate_changing}
\end{figure}

\begin{figure}[!h]
 \centering
\includegraphics[width=6.7cm, height = 5.3cm]{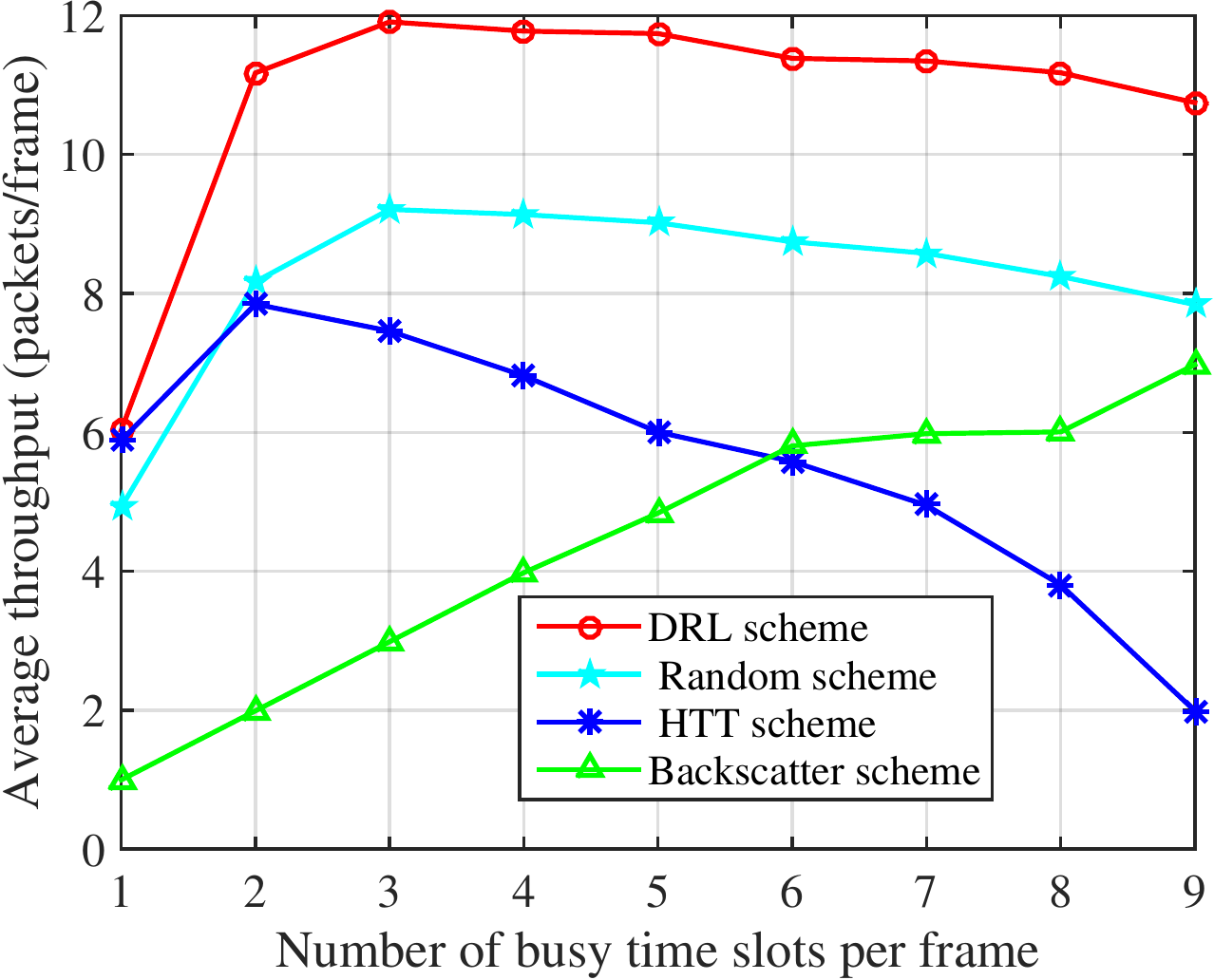}
 \caption{Average throughput versus the number of busy time slots.}
  \label{busy_time_slot_changing}
\end{figure}

To evaluate the performance of the proposed DRL algorithm, we consider different scenarios by varying the number of busy time slots per time frame, i.e., by varying $\tau$, and the packet arrival probability $\lambda$. The simulation results for the throughput versus episode are shown in Fig.~ \ref{throughput_comparison}, those for the throughput versus the packet arrival probability are illustrated in Fig.~\ref{data_rate_changing}, and those for the throughput versus the number of busy time slots are provided in Fig.~\ref{busy_time_slot_changing}.

\begin{figure}[h]
 \centering
\includegraphics[width=\linewidth]{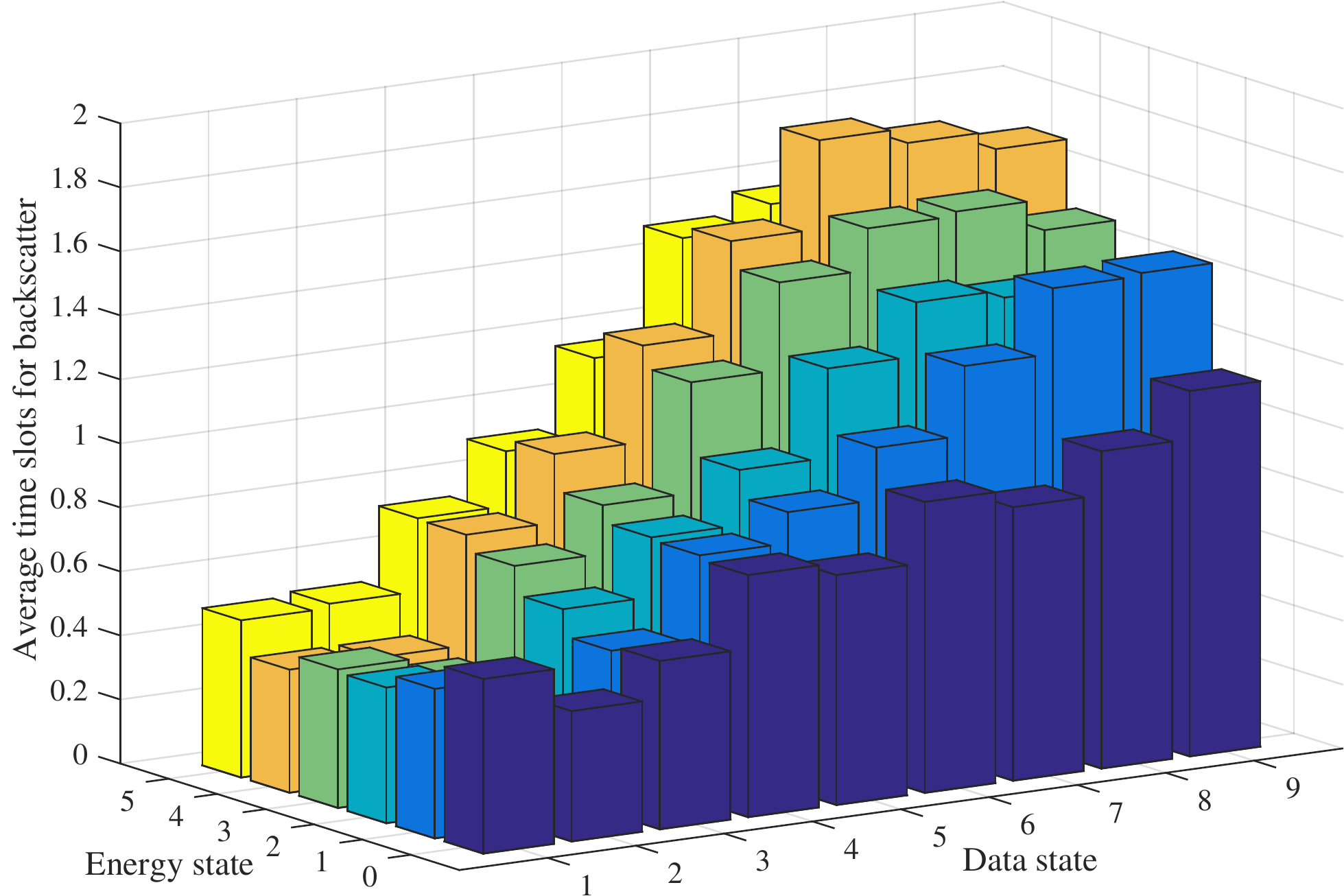}
 \caption{Average time assigned for backscatter of secondary transmitter 1.}
  \label{time_for_backscatter_actions}
\end{figure}

\begin{figure}[h]
 \centering
\includegraphics[width=\linewidth]{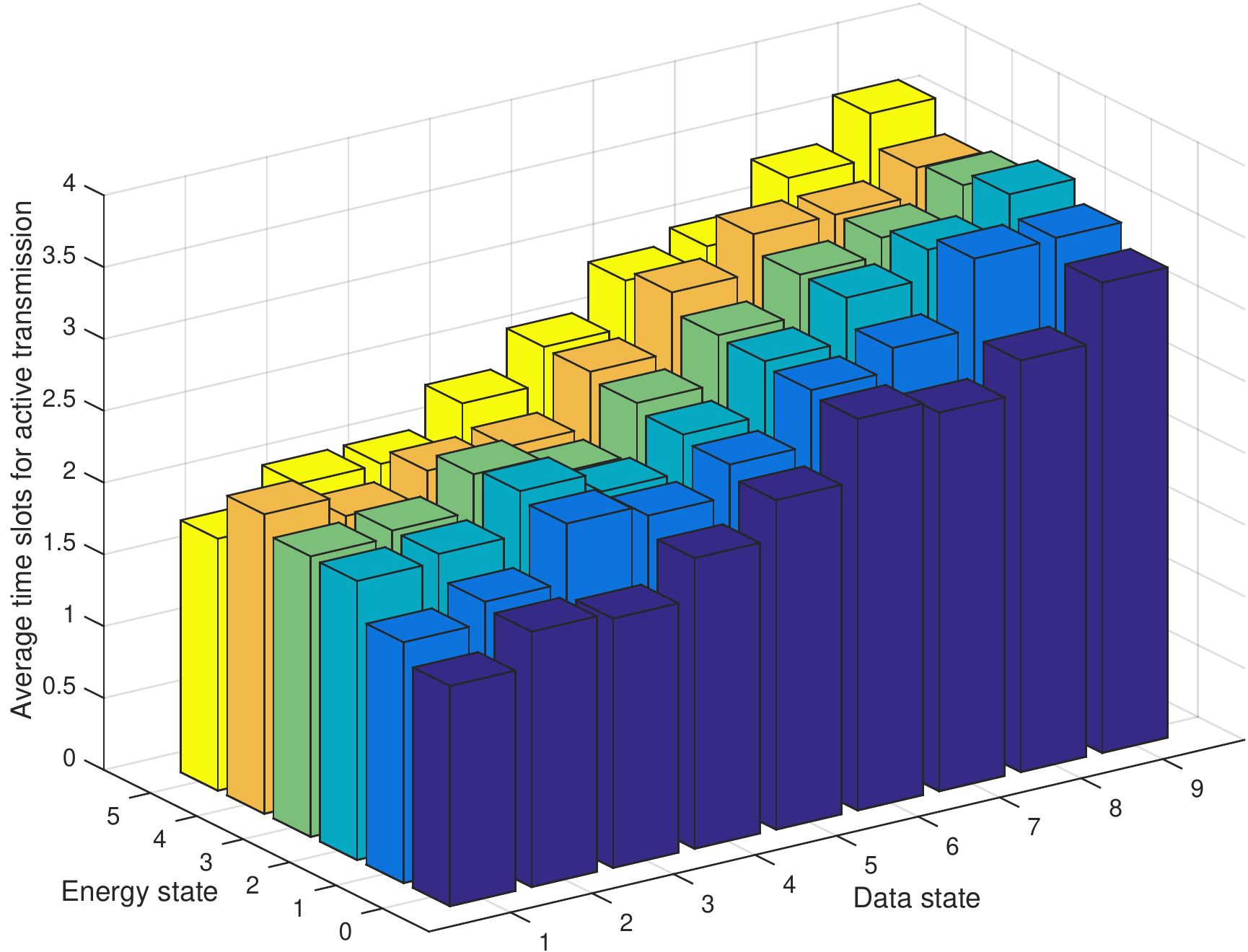}
 \caption{Average time assigned for active transmission of secondary transmitter 1.}
  \label{time_for_transmit_actions}
\end{figure}

Note that the throughput is the sum of the number of packets successfully transmitted by all secondary transmitters. In particular for the proposed DRL algorithm, the throughput depends heavily on the time scheduling policy of the gateway. This means that to achieve the high throughput, the gateway needs to take proper actions, e.g., assigning the number of time slots to the secondary transmitters for the data backscatter, data transmission, and energy harvesting. Thus, it is worth to consider how the gateway takes the optimal actions for each secondary transmitter given its state. Without loss of generality, we consider the average number of time slots that the gateway assigns to secondary transmitter 1 for the data backscatter (Fig.~\ref{time_for_backscatter_actions}) and the data transmission (Fig.~\ref{time_for_transmit_actions}). From Fig.~\ref{time_for_backscatter_actions}, the average number of time slots assigned to secondary transmitter 1 for the backscatter increases as its data queue increases. The reason is that as the data queue is large, the secondary transmitter needs more time slots to backcastter its packets. Thus, the gateway assigns more time slots to the secondary transmitter to maximize the throughput. It is also seen from the figure that the average number of time slots assigned to the secondary transmitter 1 for the backscatter increases as its energy state increases. The reason is that as the energy state of the secondary transmitter is already high, the gateway assigns less time slots for the energy harvesting and prioritizes more time slots for the backscatter to improve the network throughput. 

The secondary transmitter with a high energy state can transmit more packets in the active transmission. However, to transmit more packets, the gateway should assigns more time slots to the secondary transmitter. As illustrated in Fig.~\ref{time_for_transmit_actions}, by using the DRL algorithm, the average number of time slots assigned to secondary transmitter 1 increases as its energy state increases. 

The above results show that the proposed DRL algorithm enables the gateway to learn actions so as to improve the network throughput. As shown in Fig.~\ref{throughput_comparison}, after the learning time of around $2000$ episodes, the proposed DRL algorithm converges to an average throughput which is much higher than that of the baseline schemes. In particular, the average throughput obtained by the proposed DRL scheme is around $12$ packets per frame while those obtained by the random scheme, HTT scheme, and backscatter scheme are $9$, $7.5$, and $3$ packets per frame, respectively.  

The performance improvement of the proposed DRL scheme compared with the baseline schemes is maintained when varying the packet arrival probability and the number of busy time slots in the frame. In particular, as shown in Fig.~\ref{data_rate_changing}, the average throughput obtained by the proposed DRL scheme is significantly higher than those obtained by the baseline schemes. For example, given a packet arrival probability of $0.6$, the average throughput obtained by the proposed DRL scheme is around $15$ packets per frame while those of the random scheme, HTT scheme, and backscatter communication scheme respectively are $10$, $9.3$, and $3$ packets per frame. The gap between the proposed DRL scheme and the baseline schemes becomes larger as the packet arrival probability increases. The throughput improvement is clearly achieved as the number of busy time slots varies as shown in Fig.~\ref{busy_time_slot_changing}.

In summary, the simulation results shown in this section
confirm that the DRL algorithm is able to solve the computation expensive problem of the
large action and state spaces of the Q-learning. Also, the proposed DRL algorithm can be used for the gateway to learn the optimal policy. The policy allows the gateway to assign optimally time slots to the secondary transmitters for the energy harvesting, data backscatter, and data transmission to maximize the network throughput. 

\section{Conclusions}
In this paper, we have presented the DRL algorithm for
the time scheduling in the RF-powered backscatter cognitive radio network. Specifically, we have formulated the time scheduling of the secondary gateway as a stochastic optimization problem. To solve the problem, we have developed a DRL algorithm using DDQN including the online and target networks. The simulation results show that the proposed DRL
algorithm enables the gateway to learn an optimal time scheduling policy which 
maximizes the network throughput. The throughput obtained by the proposed DRL algorithm is significantly higher than those of the non-learning algorithms.


\end{document}